\documentclass[sn-mathphys,Numbered]{sn-jnl}

\usepackage{graphicx}%
\usepackage{multirow}%
\usepackage{amsmath,amssymb,amsfonts}%
\usepackage{amsthm}%
\usepackage{mathrsfs}%
\usepackage[title]{appendix}%
\usepackage{xcolor}%
\usepackage{textcomp}%
\usepackage{manyfoot}%
\usepackage{booktabs}%
\usepackage{algorithm}%
\usepackage{algorithmicx}%
\usepackage[noend]{algpseudocode}%
\usepackage{listings}%

\begin{document}

\title[Article Title]{Assisted Path Planning for a UGV-UAV
Team Through a Stochastic Network}


\author*[1]{\fnm{Abhay Singh} \sur{Bhadoriya}}\email{abhay.singh@tamu.edu}

\author[1]{\fnm{Sivakumar} \sur{Rathinam}}\email{srathinam@tamu.edu}

\author[1]{\fnm{Swaroop} \sur{Darbha}}\email{dswaroop@tamu.edu}

\author[2]{\fnm{David W.} \sur{Casbeer}}\email{david.casbeer@us.af.mil}

\author[3]{\fnm{Satyanarayana G.} \sur{Manyam}}\email{smanyam@infoscitex.com}

\affil*[1]{\orgdiv{Department of Mechanical Engineering}, \orgname{Texas A\&M University}, \orgaddress{\city{College Station}, \postcode{77843}, \state{TX}, \country{USA}}}

\affil[2]{\orgdiv{Controls Center}, \orgname{Air Force Research Laboratory}, \orgaddress{\city{Wright-Patterson AFB}, \postcode{45433}, \state{OH}, \country{USA}}}

\affil[3]{\orgname{Infoscitex Corporation, a DCS Company}, \orgaddress{\city{Dayton}, \postcode{45431}, \state{OH}, \country{USA}}}


\abstract{
In this article, we consider a multi-agent path planning problem in a stochastic environment. The environment, which can be an urban road network, is represented by a graph where the travel time for selected road segments (impeded edges) is a random variable because of traffic congestion. An unmanned ground vehicle (UGV) wishes to travel from a starting location to a destination while minimizing the arrival time at the destination. UGV can traverse through an impeded edge but the true travel time is only realized at the end of that edge. This implies that the UGV can potentially get stuck in an impeded edge with high travel time. 
A support vehicle, such as a unmanned aerial vehicle (UAV) is simultaneously deployed from its starting position to assist the UGV by inspecting and realizing the true cost of impeded edges. With the updated information from UAV, UGV can efficiently reroute its path to the destination.
The UGV do not wait at any time until it reaches the destination. The UAV is permitted to terminate its path at any vertex.
The goal is then to develop an online algorithm to determine efficient paths for the UGV and the UAV based on the current information so that the UGV reaches the destination in minimum time. We refer to this problem as Stochastic Assisted Path Planning (SAPP). We present Dynamic $k$-Shortest Path Planning (D*KSPP) algorithm for the UGV planning and Rural Postman Problem (RPP) formulation for the UAV planning. Due to the scalability challenges of RPP, we also present a heuristic based Priority Assignment Algorithm (PAA) for the UAV planning. Computational results are presented to corroborate the effectiveness of the proposed algorithm to solve SAPP.
}




\maketitle

\section{Introduction}\label{Intro}

Multi-agent path planning is a growing field of research with diverse applications, offering solutions for challenging tasks where multiple agents navigate through shared environments. These problems can be further subdivided into two major categories. One, the agents are trying to achieve their independent objectives while sharing given resources, also known as cooperative path planning \cite{sharon2015conflict, wagner2015subdimensional, de2013push}. Second, the agents are working together to achieve a common objective, also known as collaborative path planning \cite{dalmasso2021human, parker_thesis, donald_box_pushing}. These two categories can certainly have some overlap. In this article, we focus on a specific collaborative path-planning problem, called assisted path planning where a secondary agent is used to assist the primary agent in achieving its objective. A few applications of assisted path planning include escort missions with emergency vehicles such as an ambulance or a firetruck, search-and-rescue missions conducted during natural disasters like hurricanes, flooding, or wildfires, and others \cite{li2016hybrid, garcia2017coordinated, kitano1999robocup, berger2015innovative}. Previously, we considered assisted path planning in a partially impeded environment, where the primary agent is trying to reach its destination through a given network, and the secondary agent is trying to assist by clearing the obstructed path for the primary agent \cite{bhadoriya2023optimal}. It is important to note that the secondary agent does not have a destination and is only there to help the primary agent. In the previous work, it was assumed that the cost of clearing an obstruction is a known quantity for both agents. However, in certain applications, it may not be possible to know the exact cost of clearing an obstruction a priori. Furthermore, in specific applications, the characteristics of the obstruction may be such that it cannot be removed and therefore it is beneficial for the primary agent to re-route its path to the destination. 

In this paper, we focus on path planning for two autonomous agents operating in a stochastic environment to achieve a common objective. As an example, this environment can be a road network in an urban area where the junctions represent nodes and a road connecting two junctions represents the corresponding edge in a graph. The travel time between the two junctions can be unknown due to traffic congestion. We assume that the selected set of edges affected by traffic are known to us and are referred to as impeded edges. The travel time (or cost) of an impeded edge is considered to be a random variable whose \textit{true} value depends on the severity of the traffic congestion. 
The primary agent, an unmanned ground vehicle (UGV) in this case, must start from a specified location and reach a given destination while minimizing the arrival time at the destination. The UGV is allowed to traverse through an impeded edge and can only realize the true cost of an impeded edge when it reaches the other end after completely traversing through that edge. Therefore, UGV can get stuck in an edge for a long time if the actual cost is significantly higher compared to the expected cost. Here we assume that if UGV decides to take an edge, it cannot revert back from the middle of the edge. We also assume that UGV does not pause/wait at any time until it reaches the destination. Both these assumptions align with the behavior of a vehicle moving with the traffic flow.  
The secondary agent, an unmanned aerial vehicle (UAV) in this case, can be deployed simultaneously to assist the UGV by inspecting the traffic of the impeded edges in advance and conveying the information to the UGV so that it can re-route its path if needed. We assume that the UAV can only realize the true cost of an impeded edge by completely inspecting it, in essence, by moving along the edge from one end to the other. Once the true cost of an impeded edge is realized, it is assumed to remain the same. The path of the UAV is assumed to be unobstructed and has a known travel time.
In this particular example, the UAV can travel between any two nodes in the graph as it is not constrained by the road network. Therefore, in more general scenarios, we can have different sets of edges for the primary and secondary agents while they share the same set of nodes.

In this problem, the UGV needs to decide the path that would lead to the destination in minimum time and the UAV needs to decide on the set of edges it should inspect and the order of inspection that would be most helpful to the UGV. These decisions for both vehicles are tightly coupled and complex in nature. For example, the path for the UGV will depend on which impeded edges have already been inspected and could be inspected by the UAV before the UGV reaches them. On the other hand, the UAV will try to inspect the impeded edges on the UGV path before the UGV reaches them. But at the same time, the UAV may also want to inspect some other impeded edges in the neighborhood along its way, which might be helpful during re-routing for UGV, without losing too much time. In essence, the UAV will try to gather as much information as possible without compromising the UGV. We refer to this problem as Stochastic Assisted Path Planning (SAPP).
Given the time-sensitivity of related applications, such as emergency vehicle escort missions where there is not enough time for offline planning, we focus on developing an online algorithm for SAPP to generate high-quality, feasible solutions in real-time. 

In recent years, we have seen extensive research on problems similar to SAPP. \cite{sundar2017path} addresses a path planning problem with heterogeneous vehicles with uncertain service times for each vehicle-target pair. The study introduces a two-stage stochastic formulation to effectively solve the problem. \cite{chirala2023uv} considers a multi-agent mission planning problem under uncertainty with vehicle availability and presents a skewed variable neighborhood search algorithm to solve large instances. \cite{fu2022robust} presents a multi-agent task scheduling problem with the uncertainty in agent's capability.
The deterministic version of assisted path planning, referred to as Assisted Shortest Path Planning (ASPP), has been studied in \cite{bhadoriya2023optimal}, where the impeded edges have known obstructions and the cost to clear those obstructions is also known \textit{a priori}. The authors have presented a generalized permanent labeling algorithm (GPLA*) to find the problem optimally. \cite{bhadoriya2023multi} present a Monte-Carlo Tree Search (MCTS) based anytime algorithm to solve ASPP for time-sensitive applications.

The main contribution of this research is to develop an online algorithm to efficiently solve the SAPP in real-time. We first present a strategy that breaks the SAPP into two sub-problems of path planning for the two agents. Then, we propose a dynamic $k$-shortest path planning algorithm (D*KSPP) to find and update $k$-shortest paths for the UGV to reach the destination. D*KSPP is a hybrid algorithm that combines D* Lite \cite{koenig2002d} and Yen's algorithm \cite{yen1971finding} to efficiently update the $k$-shortest paths for the UGV when new information is available. For UAV path planning, we propose a rural postman problem with time windows (RPP-TW) \cite{monroy2014rural} formulation. Due to the scaling challenges of RPP-TW, we also propose a heuristic-based priority assignment algorithm (PAA) that performs as well as RPP-TW with a fraction of computation time.

The structure of this paper is as follows. Section \ref{prob_state} presents the formal description of SAPP. Section \ref{Algo} presents the solution strategy and all the proposed algorithms. Section \ref{comp_study} presents the computational study of the proposed algorithms. We also compare our algorithm against a naive approach and discuss the effectiveness of the proposed algorithms for complex graph structures. Further, the limitations of all proposed algorithms are presented. In the end, we present a real-world case study from six different cities in the United States. Section \ref{conclusion} presents concluding remarks and future work.

\section{Problem Statement} \label{prob_state}
This section introduces the required notation to define the SAPP. Let $G=(V, E, S)$ be an undirected, connected graph representing a stochastic environment. $V$ is the set of vertices in the environment, $E$ is the set of undirected edges for the UGV and $S$ is the set of undirected edges for the UAV where $S\supseteq E$. The UGV and the UAV start at vertices $p$ and $q$, respectively, where $p$ and $q$ need not be identical. Let $d \in V$ be the destination of the UGV. The UAV may terminate at any vertex at any time and therefore has no destination. For the remainder of this paper, we will use cost and time interchangeably. Let $K\subset E$ be the set of edges that can be potentially obstructed and are referred to as \textit{impeded edges}. The remaining edges are called \textit{unimpeded edges}. 
Each unimpeded edge $e\in E\setminus K$ is assigned a positive travel cost $T_e$ for the UGV. Each edge $e\in S$ is assigned a positive travel cost $\tau_e$ for the UAV. For each edge $e\in K$, the travel cost for the UGV is a non-negative continuous random variable represented by $\mathcal{T}_e$ with a known distribution. We assume the domain of $\mathcal{T}_e$ is known and ranges from $[T_e^{min}, T_e^{max}]$. We will say the true cost of an impeded edge $e$ is realized when either the UAV or UGV completely observes that edge i.e. when they reach the other end of the edge after traversing through it. The true cost is always a positive finite value and once it is realized, it will remain the same.
We will make the following assumptions: (1) both vehicles start at the same time, (2) the two vehicles communicate at all times and information is shared in a negligible amount of time, (3) UGV will not wait/pause at any time, (4) both vehicles will not change their direction or terminate in the middle of an edge traversal. 

After both vehicles begin from their starting location, they move according to their respective strategies until the UGV reaches the destination. The UGV will choose the best possible path to the destination based on the available information and the UAV will decide on which impeded edges should be inspected and in what order to best serve the UGV. The UGV is allowed to update its path to the destination as soon as new information is available and consequently, UAV can update its path as well. We assume the rerouting computation time to be negligible for both vehicles compared to their travel time. If the UGV arrives at the end of an edge $e\in K$ that lies on its path and whose true cost is not known yet, it will continue on its path without pausing and traverse the edge. In this scenario, the UGV will incur the true cost of that edge and therefore no longer need the UAV to inspect it.
The objective of SAPP is then to minimize the time of arrival of the UGV at the destination $d$.

\begin{figure}[htbp]
\begin{center}
\includegraphics[width=\linewidth]{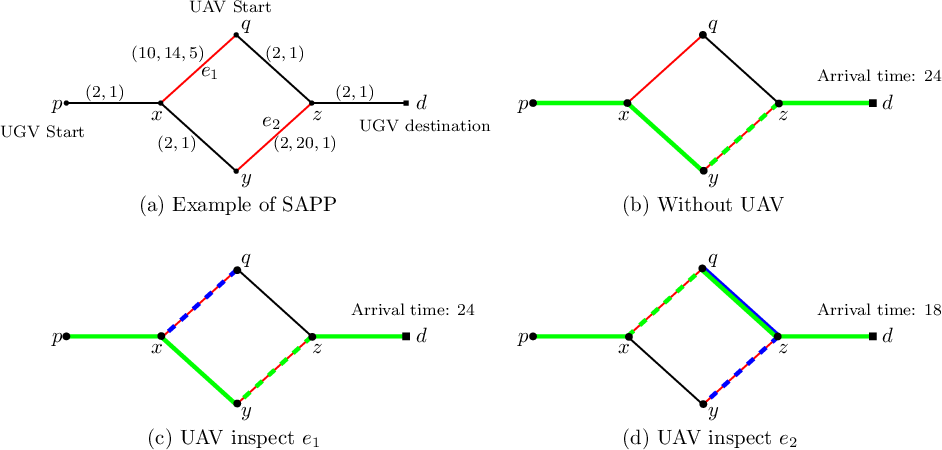}
\caption{(a) A sample instance for the SAPP. Black edges represent unimpeded edges and red edges represent impeded edges. UAV has the same set of edges as UGV. Travel times are represented as follows: for unimpeded edges $(T_e, \tau_e)$, for impeded edges $(T_e^{min}, T_e^{max}, \tau_e)$. Let the true cost of $e_1$ and $e_2$ be 12 and 18.
In (b), (c), and (d) the convoy and the service vehicle paths are represented by green and blue colors, respectively. The dashed path represents the true cost of the impeded edge realized by the vehicle associated with the corresponding color. In (b), the UGV reaches the destination without the assistance of the UAV. In (c), the UAV decides to inspect the nearby edge $e_1$, whereas in (d), the UAV inspects the edge $e_2$.
}
\label{SAPPinstance} 
\end{center}
\end{figure}

A sample problem of SAPP is shown in Figure \ref{SAPPinstance}(a). In this example, impeded edges $e_1$ and $e_2$ are highlighted with red color. Here we assume that the UAV has the same set of edges as the UGV i.e. $S=E$. UGV and UAV travel times are indicated using tuples $(T_e, \tau_e)$ for each unimpeded edge. For each impeded edge, we assign a tuple $(T_e^{min}, T_e^{max}, \tau_e)$ to indicate the minimum and maximum travel time for the UGV and the travel time for the UAV. We assume a uniform distribution for the cost of impeded edges therefore, the expected cost for $e_1$ and $e_2$ are 12 and 11. For this example, let the true cost for $e_1$ and $e_2$ be 12 and 18 respectively. Figure \ref{SAPPinstance}(b) represents the scenario where UAV support is not available. In this case, UGV takes the path based on the expected cost and the final arrival time at the destination is 24. 
Figure \ref{SAPPinstance}(c) represents the scenario where the UAV first inspects the impeded edge $e_1$ as it is nearby. As the UGV is following the path based on the expected cost, it will be on the edge $e_2$ when the UAV finishes inspecting edge $e_1$ and realizes the cost. Therefore, the contribution of the UAV is irrelevant in this case.
Figure \ref{SAPPinstance}(d) represents the scenario where the UAV decides to first inspect the impeded edge $e_2$. The UAV will realize the true cost of $e_2$ when the UGV reaches node $x$. As the true cost of $e_2$ is 18, the UGV will then reroute its path via node $q$ to $d$. This results in the final arrival time of 18.


\section{Algorithm} \label{Algo}
As discussed before, the decisions made by both vehicles are tightly coupled and therefore it is computationally expensive to solve SAPP using any deterministic approach. 
This paper focuses on developing an online algorithm for the SAPP that is time-efficient and provides a high-quality solution. 
We divide the algorithm into the following steps: (1) find the best possible path for the UGV to reach the destination based on current information, (2) identify the best set of impeded edges that should be inspected by the UAV based on the graph structure and UGV's current path. We refer to these edges as \textit{critical edges}, (3) find a path for the UAV to inspect critical edges efficiently and (4) repeat the previous steps when the true cost of an impeded edge is realized until the UGV reaches its destination.
It is important to note that all the above steps should be computationally very fast for an online algorithm. Therefore, we want to come up with simple yet effective strategies for re-planning. For the UGV, we would like to update its path from its current location to the destination every time we have new information. D* Lite \cite{koenig2002d} is the most efficient algorithm for re-computing the shortest path of an agent whenever the cost of an edge gets updated in the graph.

For the UAV, we propose the following approach to identify the critical edges for inspection. First, find the $k$-shortest paths for the UGV from its current position to the destination. All the impeded edges with unrealized costs that lie on these $k$ paths are added to the set of critical edges. With this approach, we not only consider the impeded edges on the current path of UGV but also keep other impeded edges in consideration which lie on alternative paths for the convoy. Also, we can control our area of interest by adjusting the $k$ value. After identifying the critical edges, we need to find an efficient path for the UAV to cover these edges before the UGV can reach them. This can be formulated as a Rural Postman Problem with Time Window (RPP-TW). Note here that in any path planning algorithm for UGV (D* or $k$-shortest path), we will need some cost estimate for the impeded edges. Therefore, we will use the expected cost of the impeded edges during the planning for UGV. Also, we have combined the ideas from the D* Lite and $k$-shortest path planning algorithms to get the updated $k$ shortest paths after the cost of an impeded edge is realized. We refer to this algorithm as the Dynamic $k$-Shortest Path Planning (D*-KSPP) algorithm. Next, we will present these algorithms in more detail.

\subsection{Dynamic K-Shortest Path Planning Algorithm (D*-KSPP)} \label{KSPP}

In this section, we will first briefly summarize the D* lite algorithm and Yen's algorithm \cite{yen1971finding} for $k$-shortest paths and then present the algorithm for D*-KSPP.
D* lite is used to efficiently re-route the path of the agent to the destination after some of the edges in the graph experience a change in their travel cost. The key idea behind this algorithm is to use the information from the previous search and focus the next search on the relevant part of the graph.
In D* lite, we perform the search starting from the destination because the location of UGV changes as it moves but the destination is fixed. Therefore we keep a record of the distance of a node from the destination. D* lite maintains two estimates, $g(v)$ and $rhs(v)$ for each vertex $v\in V$. $g(v)$ estimates the distance from the destination for the vertex $v$ and is directly related to the g-values of an A* search. $rhs(v)$ also provides an estimate of the distance to the destination but it is a one-step look-ahead value based on the $g$ values of the neighbors of $v$ i.e. $N(v)$. 

\begin{equation}
    rhs(v) =  
    \begin{cases}
    0,              & \text{if } v=d\\
    min_{v'\in N(v)} (g(v')+c(v',v)) ,& \text{otherwise} 
\end{cases}
\end{equation}

A vertex is called locally consistent if its $g$-value equals its $rhs$-value, otherwise, it is called locally inconsistent. We also maintain a priority queue $Q$ same as the A* algorithm, however, the priority key $k(v)$ of each node has two elements here.

\begin{align*}
    k(v)   &= [k_1(v); k_2(v)]\\
    k_1(v) &= min(g(v),rhs(v)) + h(v, v_{curr}) + k_m\\
    k_2(c) &= min(g(v), rhs(v))
\end{align*}
\noindent
$v_{curr}$ refers to the current position of the vehicle. $h$ is any consistent heuristic i.e. it is admissible and follows the triangle inequality. $k_1$ and $k_2$ correspond to the $f$-value and the $g$-value used in A*. In the priority queue, we sort the keys in increasing order of $k_1$, and in case of a tie, we pick the key with smaller $k_2$. Further ties can be broken arbitrarily.
%
\begin{algorithm}
        \caption{D* lite \label{alg:D*}}
        \begin{algorithmic}[1]

            \vspace{2mm}

            \Procedure{D*lite}{$v_{curr}$, $g$, $rhs$, Q, $I$, $k_m$}        \Comment{$I$ : edges with updated cost} 
                \For{ $(u,v), c_{new} \in$ $I$}
                    \State $rhs$Update($(u,v), c_{new}$)
                \EndFor
                \State ComputeShortestPath()
                \State \textbf{return} path($v_{curr}, d$)
                
            \EndProcedure

            \vspace{2mm}
            
            \Procedure{$rhs$Update}{$(u,v), c_{new}$}
                \If{$c_{old}>c_{new}$}
                    \State $rhs(u) \gets min(rhs(u), g(v) + c_{new})$
                    \State $rhs(v) \gets min(rhs(v), g(u) + c_{new})$
                \Else{}
                    \If{$rhs(u) = g(v)+c_{old}$ AND $u\neq d$}
                        \State $rhs(u) \gets min_{s'\in N(u)}(rhs(u), g(s') + c(s', u))$
                    \EndIf
                    \If{$rhs(v) = g(u) + c_{old}$ AND $v\neq d$}
                        \State $rhs(v) \gets min_{s'\in N(v)}(rhs(v), g(s') + c(s', v))$
                    \EndIf
                \EndIf
                \State UpdateVertex($u$)    
                \State UpdateVertex($v$)
            \EndProcedure

            \vspace{2mm}

            \Procedure{ComputeShortestPath}{}   \Comment{Appendix \ref{AppA}}
            \EndProcedure
            
            \vspace{2mm}

            \Procedure{UpdateVertex}{$v$}     \Comment{Appendix \ref{AppA}}
                
             
            \EndProcedure

            \vspace{2mm}

            \Procedure{CalculateKey}{$v$} \Comment{Appendix \ref{AppA}}
            \EndProcedure

            \vspace{2mm}
            
        \end{algorithmic}
\end{algorithm}

$k_m$ is used to maintain the order in the priority queue between the searches as the $v_{curr}$ changes. When D* lite is initialized, it computes the shortest path to the destination exactly the same as A*. Then, as the cost of an edge is changed, we update the $rhs$ value of all affected vertices (beyond the immediate nodes of the affected edge) and add them to the priority queue if needed with the updated key value. Next, we iteratively expand the nodes from the priority queue with the minimum key value until the $v_{curr}$ becomes consistent or all nodes in the priority queue have a key value larger than the $k(v_{curr})$. This approach utilizes the $g$ and $rhs$ values from the previous search and only considers the inconsistent nodes after the edge cost update. We refer the reader to \cite{koenig2002d} for more detail on the D* lite algorithm.

\begin{algorithm}
        \caption{Dynamic $k$-shortest path planning algorithm (D*-KSPP) \label{alg:DKSPP}}
        \begin{algorithmic}[1]
            
            \Procedure{Initialize}{}
                \State Q = $\emptyset$; $v_{old} = p$; $k_m = 0$
                \For{$v\in V$} $rhs(v) = g(v) = \infty$
                \EndFor
                \State $rhs(d) = 0$
                \State Q.Insert($d, [h(p, d); 0]$)
            \EndProcedure

            \vspace{2mm}
            
            \Procedure{Update $k$-paths}{$v_{curr}$, $I$} \Comment{$I$ : edges with updated cost} 
            \State $A = B = \emptyset$
            \State $k_m \gets k_m + h(v_{old}, v_{curr})$
            \State $A^1\gets$ D*lite($v_{curr}$, $g$, $rhs$, Q, $I$, $k_m$)
            
            \For{$k$ from 2 to k}
                \For{$i$ from 1 to $|A^{k-1}|-1$}
                    \State $R^k_i, s^k_i \gets$ getRootPath, getSpurNode
                    \State $E_i^k \gets$ remove unwanted edges 
                    \State $g'$, $rhs'$, Q$'$ $\gets$ $g$.copy(), $rhs$.copy(), Q.copy()
                    \State $k_m' \gets k_m + h(v_{curr}, s_i^k)$
                    \State $S_i^k\gets$ D*lite($s^k_i$, $g'$, $rhs'$, Q$'$, $E_i^k$, $k_m'$)
                    \State $A_i^k \gets R_i^k+S_i^k$
                    \If{$A_i^k$ not in $B$}
                        $B$.add($A_i^k$)
                    \EndIf
                    \State restore $E_i^k$ to Graph
                \EndFor
                
                \If{$B$ is empty} break
                \EndIf
                \State $B$.sort
                \State $A^k\gets B[0]$ 
            \EndFor
            \State$v_{old} = v_{curr}$
            \State \textbf{return} $A$
            \EndProcedure
        \end{algorithmic}
\end{algorithm}

Next, we will discuss Yen's algorithm to compute the $k$-shortest paths for the UGV. We assume two containers, $A$ and $B$, where $A$ will hold the $k$-shortest paths and $B$ will hold all the potential paths that can be added to $A$. Let $A^k$ represent the $k^{th}$ shortest path and $R^k_i$ be a sub-path consisting of the first $i$ nodes of $A^{k-1}$ and referred a root path of $A^k$. The last node of $R^k_i$ is also referred to as the spur node, $s^k_i$.
We start with finding the shortest path using any existing method and set it to $A^1$. In order to find the $A^k$, $A^1, A^2, ... , A^{k-1}$ must be known. Next, we divide the iteration to find the $k^{th}$ shortest path in the following steps.
\begin{enumerate}
    \item For each $i = 1, 2, ..., len(A^{k-1})-1$, check if $R^k_i$ coincides with the sub-path consisting of the first $i$ nodes of $A^j$ in sequence for $j = 1, 2, ..., k-1$. If so, remove the edge between the node $i$ and $i+1$ in the path $A^j$ i.e. set the cost to infinity. Note that this change is only for this iteration and the edge cost is restored to its original value at the end of this iteration.
    \item Remove all the nodes in $R^k_i$ from the graph except for the spur node $s^k_i$. We will restore these nodes after each iteration.
    \item Find the shortest path from $s^k_i$ to the destination. This sub-path is referred to as spur path $S^k_i$. Find the new path by joining $R^k_i$ and $S^k_i$ and add it to list $B$.
    \item Find the path(s) with the minimum length in $B$ and shift them to $A$.
\end{enumerate}

We repeat the above steps until we get the $k$ paths in $A$ or $B$ is empty. Note that in step 2, we want to remove certain nodes from the graph, which is equivalent to setting the cost of all edges incident on to that node to infinity. Note that Yen's algorithm results in simple paths, i.e. they do not contain a cycle.

Next, we present the overall structure of the D*-KSPP and the methodology to combine the D* lite and Yen's algorithms. Algorithm \ref{alg:D*} presents the methods in the D* lite algorithm and Algorithm \ref{alg:DKSPP} gives the overview of D*-KSPP. In Algorithm $\ref{alg:D*}$, CaluculateKey, UpdateVertex, and ComputeShortestPath methods are the same as presented in \cite{koenig2002d}. We refer the reader to Appendix \ref{AppA} for more details on these methods. We have introduced two new methods to facilitate the integration with D*-KSPP. Given an edge and its updated cost, $rhs$Update is used to update the $rhs$-value of the vertices associated with that edge. D* Lite method takes in the current vertex $v_{curr}$, $g$, $rhs$, $Q$, set of edges with updated cost, and $k_m$ and returns the shortest path from $v_{curr}$ to $d$. Note that $v_{curr}$ need not be the position of the UGV as we will discuss later.

We initialize the D*-KSPP by setting the $g$ and $rhs$ values to infinity for all nodes except for the destination $d$ whose $rhs$ value is set to zero and insert the vertex $d$ in the priority queue $Q$. We maintain the variables $g$, $rhs$, and $Q$ throughout the algorithm until the UGV reaches the destination. We call the Update $k$-Paths method every time we have new information. It takes the UGV's current position and the impeded edges with the updated cost as an argument and returns updated $k$-shortest paths. We start with finding the shortest path from $v_{curr}$ to $d$ using D* lite and assign it as $A^1$. Note that, in this step, $g$, $rhs$, and $Q$ variables get updated according to Algorithm \ref{alg:D*}. From lines 10 to 22, we follow Yen's algorithm to find $A^k$. Line 13 represents steps 1 and 2 of Yen's algorithm that sets the cost of certain edges to infinity. In line 16, we again use the D* lite to find the spur path $S_i^k$. Here we pass the spur node $s_i^k$ to the D* lite algorithm, which is not the current position of the UGV.
Also note that we pass a copy of $g$, $rhs$, and $Q$ to D* lite. This is done to avoid modifying the original variables as they are independent of the spur path computation. We utilize the $g$ and $rhs$ values from the previous search of computing $A^1$ to efficiently find spur paths. This is a key step behind the performance of the D*-KSPP as we may need to find $\mathcal{O}(k|V|)$ spur paths in each replanning step. The remaining steps follow Yen's algorithm and return $A$ with $k$-shortest paths for the UGV in the end. Next, we will discuss the path planning for the UAV.

\subsection{Rural Postman Problem with Time Window (RPP-TW)} \label{RPP}
In this section, we will first present a method to setup the UAV path planning as a Rural Postmen Problem with Time Window constraints and then propose a solution strategy. As discussed in section \ref{Algo}, we will first identify the set of critical edges $E_c$ to inspect from the $k$-shortest path of the UGV. For each edge $e\in E_c$, we assign a corresponding time window $[t_{min}, t_{max}]$ within which it should be visited. For our problem, we can set $t_{min} = 0$ for all $e\in E_c$ as we want the UAV to visit these edges as early as possible. For all $e\in E_c$ that lie on $A^1$, we set $T_{max}$ to the minimum arrival time of the UGV at $e$. The minimum arrival time can be computed using the minimum cost for the impeded edges. For all other edges in $E_c$ that do not lie on $A^1$, set $T_{max}$ to infinity as the UGV is not on those corresponding paths. Also, note that all $E_c\subset S$ as $E_c \subset E \subseteq S$.

\begin{algorithm}{}
        \caption{Rural Postman Problem - Depth First Search (RPP-DFS) \label{algo:RPP-DFS}}
        \begin{algorithmic}[1]
            
            \Procedure{Initialize}{}
                \State bestCost $\gets$ $\infty$
                \State bestVisited $\gets$ \{``0''\}
                \State DFS($v$ = ``0", Cost = 0, Visited = \{``0''\})
            \EndProcedure

            \vspace{2mm}
            
            \Procedure{DFS}{$v$, Cost, Visited}
            \State removedNodes $\gets$ Visited
            \For{$i\in$ Visited}
                \If{$i \neq $ ``0" }
                    \State removeNodes $\cup$ \{$i^o$\}
                \EndIf
            \EndFor
            \State N($v$) $\gets$ N($v$)$\setminus$removedNodes

            \State TerminalState $\gets$ True

            \For{$n$ $\in$ N($v$)}
                \State newCost $\gets$ Cost + $G_o$.cost($(v,n)$)
                \If{newCost $\leq$ $n.T_{max}$} 
                    \State newVisied $\gets$ Visited $\cup$ \{$n$\}
                    \If{(newCost $<$ bestCost) OR (newVisited $\not\subset$ bestVisited )} 
                        \State DFS($n$, newCost, newVisited)
                        \State TerminalState $\gets$ False
                    \EndIf
                \EndIf
            \EndFor

            \If{TerminalState}
                \State newCost $\gets$ Cost + $G_o$.cost((v,``0"))
                \If{($|$bestVisited$|$ $<$ $|$Visited$|$) OR (bestVisited = Visited AND bestCost $>$ Cost)}
                    \State bestCost $\gets$ Cost
                    \State bestVisited $\gets$ Visited
                \EndIf
            \EndIf
            \EndProcedure
        \end{algorithmic}
\end{algorithm}
Given a graph for UAV $G=(V,S)$ and a set of critical edges $E_c$, the objective is to find a minimum cost path that traverses all the critical edges at least once within the time window.  As discussed in \cite{monroy2014rural}, we need significant graph modification to solve RPP-TW. We have used the following transformation method from\cite{monroy2014rural} to generate a new graph $G_o = (N, S_o)$. For each $e\in E_c$, we define two nodes $i, i^o$ in $G_o$, where $i,i^o\in N$ represent the two possible directions in which edge $e$ can be traversed. $S_o$ is a set of arcs where each arc connects node $i\in N$ with node $j\in N$ if they do not correspond to the same edge in $G$. The cost of an arc that connects node $i$ to node $j$ in $G_o$ is equal to the cost of the edge represented by $i$ plus the length of the shortest path in G from the final node of the edge represented by $i$ to the initial node of the edge represented by $j$, according to the traversal directions. For the time window of each node, we reduce the $T_{max}$ by the time it takes the UAV to travel that edge. This adjustment is made to ensure that the UAV can completely traverse the edge in $G$ before $T_{max}$. The adjustment to $T_{min}$ is irrelevant to this problem.
Next, we add a pseudo node ``0" to represent the current position of the UAV.  In traditional Rural Postman Problems,  the objective is to find a tour so that the vehicle can get back to its depot. But in this problem, we do not need the UAV to go back to its current position. Therefore, we have introduced the following changes to the $G_o$ construction. We add an arc from ``0" to each $i\in N$ with a cost equal to the length of the shortest path in $G$ from the current position of UAV to the initial node of the edge represented by $i$. Then we add an arc from each $i\in N$ to ``0" with a cost equal to the cost of the edge represented by $i$. This represents a zero cost to go back to the current position of UAV from any edge $e\in E_c$. Therefore a tour in $G_o$ is equivalent to a path in $G$ from the current position of UAV to the last critical edge covered on the tour. We refer the reader to \cite{monroy2014rural} for more detail on graph transformation for RPP-TW.

We understand that there may not exist a feasible solution that visits all the edges in $E_c$ within the time constraints. In such cases, we need to find the best feasible path that visits as many edges in $E_c$ as possible within the time window constraints. We present a depth-first-search algorithm to solve RPP-TW to find the best feasible path for the UAV. Algorithm \ref{algo:RPP-DFS} presents the RPP-DFS algorithm. Here we keep track of the best solution cost we found so far and the corresponding ordered set of visited nodes in $G_o$. During each step of DFS, we have the current node, cost, and the ordered set of previously visited nodes. To determine the eligible neighbors of a node $v$ that should be considered for the next step, we remove all the nodes $i$ in the visited set and their corresponding $i^o$ node which represents the same edge as $i$ in $G$ but in the opposite direction. For each eligible neighbor of $v$, we check if it can be reached within the time window. If yes, then we recursively call the DFS for the neighbor if the new solution is not dominated by the best solution found so far (Algorithm \ref{algo:RPP-DFS}, line 16). The new solution is said to be dominated by the best solution only if the new solution cost is greater than or equal to the best solution cost and all the visited edges in the new solution are contained in the best solution. The state of DFS is considered to be terminal if it does not have any eligible neighbors or all the extensions to its neighbors are dominated by the previously found best solution. At the terminal state, if the new solution has more visited nodes or has the same set of visited nodes but at a lower cost, we update the best solution found so far with the new solution. Once the RPP-DFS terminates, the path of the UAV can obtained by following the order of visited nodes in the best solution.

\subsection{Priority Assignment Algorithm (PAA)} \label{PAA}
The RPP is shown to be an NP-complete problem and it does not scale well as the number number of critical edges increases. Also, in the proposed algorithm for the SAPP, we reroute the path of UGV and UAV as soon as any new information is available i.e. immediately after the true cost of an impeded edge is realized. Therefore it is equivalent to identifying the best first edge to inspect during each rerouting step for the UAV. We propose a Priority Assignment Algorithm (PAA) that can identify the best critical edge to inspect for the UAV and can scale linearly with the increase in the number of critical edges. In PAA, we identify several key parameters to assign a priority value to each critical edge and pick the edge with the highest priority to inspect. We identify these parameters as follows:
\begin{enumerate}
    \item \textbf{Number of paths ($p_1$)}:
    One of the most important parameters is that if a critical edge acts as a bottleneck for the UGV, i.e. many shortest paths pass through the same critical edge. We aim to assign a high priority to a critical edge that lies on multiple $k$-shortest paths of the UGV. If a critical edge lies on $n$ paths out of $k$ paths, then $p_1$ for that edge is computed as $$p_1 =  \frac{n}{k} $$

    \item \textbf{Time of divergence ($p_2$)}: 
    The UGV always follow the $A^1$ path and all the other path $A^i$ for $i = 2,...,k$ will diverge from $A^1$ at a particular vertex. The time of divergence, $\lambda$, for all the critical edges on a path $A^i$ is then the expected time to reach that divergence vertex from the UGV's current position. Note that $\lambda$ can be zero as well. $\lambda$ for a critical edge on $A^1$ is set to be the expected time of arrival of the UGV at the first vertex of that edge.
    We would want to assign a higher priority to the edges with smaller divergence time as it will help us avoid situations where UGV gets stuck in a high cost impeded edge or has to retrace its path if an alternate path is shorter. To compute $p_2$, first we find minimum and maximum divergence time across all critical edges, $\lambda_{min}$ and $\lambda_{max}$. For a critical edge with time of divergence $\lambda$, $p_2$ is computed as $$p_2 = \frac{\lambda_{max} - \lambda}{\lambda_{max}- \lambda_{min}}$$
    If $\lambda_{min} = \lambda_{max}$, then we assign $p_2 = 1$. 
    
    \item \textbf{Cost variance ($p_3$)}: 
    Realizing the true cost of critical edge with high cost-variance is very important because of the following reasons. 1) If the true cost is very high compared to the expected cost, it is better to realize it early and try to avoid that edge. 2) If the true cost is very low compared to the expected cost, it may be beneficial for us to route UGV's path through that edge. Either way, a critical edge with high cost-variance should be prioritized above other edges with low cost-variance. Let $\sigma_{max}$ be the maximum variance across all critical edges, then $p_3$ for a critical edge with cost variance $\sigma$ is $$p_3 = \frac{\sigma}{\sigma_{max}}$$
    
    \item \textbf{Relative position ($p_4$)}: 
    Between two critical edges, if all the above mentioned parameters are the same, then we should prioritize the edge which is closer to the UAV's current position. The distance of a critical edge from the UAV is the shortest distance to reach either nodes on that edge. For a critical edge which is $d$ units away from the UAV, $p_4$ is computed as
    $$p_4 = 1- \frac{d}{d_{max}}$$
    where $d_{max}$ is the maximum distance of a critical edge from the UAV. If $d_{max}=0$, then we assign $p_4 = 1$
    
\end{enumerate}

We assign a weight $w_i$ for $i=1,2,3,4$ for each of the above mentioned parameters. These weights can be adjusted to increase the influence of one parameter over the other. For each critical edge, we compute all the parameters $p_i$ and then compute the priority as the weighted sum of the parameters $\sum_{i} w_i p_i$. A critical edge with the highest priority is then chosen to be inspected by the UAV. We repeat this process every time we need to reroute for the UAV.

\section{Computational Study}\label{comp_study}
All algorithms were implemented in Python 3.6 and computations were done on a MacBook Pro laptop (10-core M2-Pro processor, 16 GB RAM). For the analysis, we have used real-world urban maps of several cities generated from OpenStreetMap\cite{OpenStreetMap}. To evaluate the effectiveness of the presented algorithm, we compare our results with a lower bound and a naive method. The lower bound can be computed by finding the shortest path for the UGV assuming we know the true cost of all impeded edges. The naive method is a strategy where the UGV follows the shorted path based on the expected cost and the UAV only visits the impeded edge on the UGV's path that is closest to UGV and can be visited within the time window. These two bounds are good indicators to assess the performance of the presented algorithm. 

In the following analysis, we assume a uniform distribution for the cost of all impeded edges between their assigned range. Other distributions can be used for different applications. Also, we assume the maximum speed of a UGV to be 1 unit without the loss of generality and the UAV to be twice as fast as the UGV. The speed ratio between the vehicles does not affect the presented algorithms and can be changed based on the application.
We refer to the presented algorithm as D*KSPP-RPP when RPP-TW formulation is used for UAV path planning and D*KSPP-PAA when we use the priority assignment algorithm for the UAV.
The analysis is divided into the following sections. Firstly, we compare our D*KSPP-RPP algorithm with the naive method and analyze the performance as we change the number of paths $k$ for the UGV. Next, we examine the computation limitations of RPP and present a comparison between RPP and PAA algorithms for UAV path planning. We also analyze the computation time for D*KSPP as we increase the graph size. In the end, we present real-world analysis by using road networks from 6 different cities in the USA.

\subsection{Comparison between D*KSPP-RPP and Naive Method} \label{KSPP-Naive}

\begin{figure}[htbp]
\begin{center}
\includegraphics[width=\linewidth]{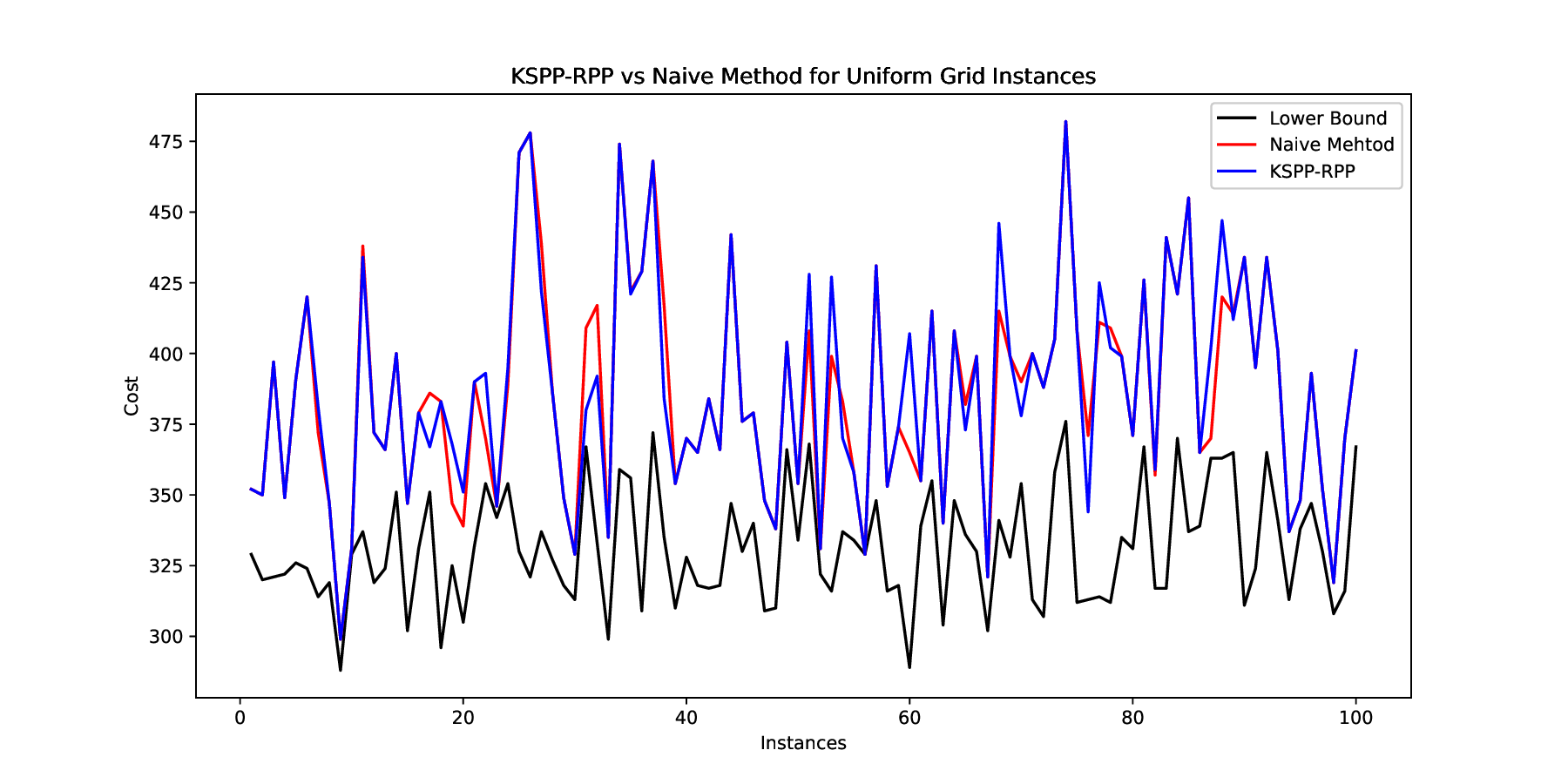}
\caption{Comparison between D*KSPP-RPP and Naive Method for 100 randomly generated grid $20\times10$ grid instances. Number of UGV paths considered for D*KSPP is 5. A lower bound is provided to indicate the quality of the solution.}
\label{fig1} 
\end{center}
\end{figure}

To compare the proposed D*KSPP-RPP algorithm with the naive method, we first consider instances with uniform grid structure. For this study, we generated 100 grid instances of size $20\times10$ on the Cartesian plane with adjacent nodes separated by $10$ units. For all of the instances, the origin and the destination of the UGV, $p$ and $d$, were chosen to be at diagonally opposite ends of the grid, $(0,0)$ and $(190,90)$ respectively, to avoid trivial cases. The starting position of the UAV was chosen randomly for each instance. For each instance, we pick 5 random cuts as the set of impeded edges $K$. $T_e$ for all unimpeded edges and $T_e^{min}$ for all impeded edges are set to the euclidean distance between the corresponding two nodes. $T_e^{max}$ is chosen randomly between $[80, \,100]$ for all $e\in K$. For UAV, the travel cost between any two nodes is equal to the euclidean distance between the nodes divided by the UAV speed.
Figure \ref{fig1} shows the performance of the D*KSPP-RPP algorithm with the number of UGV paths set to $k = 5$ and the naive method. Both algorithms are compared against the proposed lower bound. As we can see, the performance of the naive method is as good as the D*KSPP-RPP algorithm for randomly generated uniform grid instances. Moreover, there are a few instances where the naive method turns out to be a slightly better strategy than the proposed algorithm. This can happen in some specific cases where the best impeded edge to inspect lies on UGV's path and the RPP algorithm may decide to visit a nearby impeded edge first.

\begin{figure}[htbp]
\begin{center}
\includegraphics[width=\linewidth]{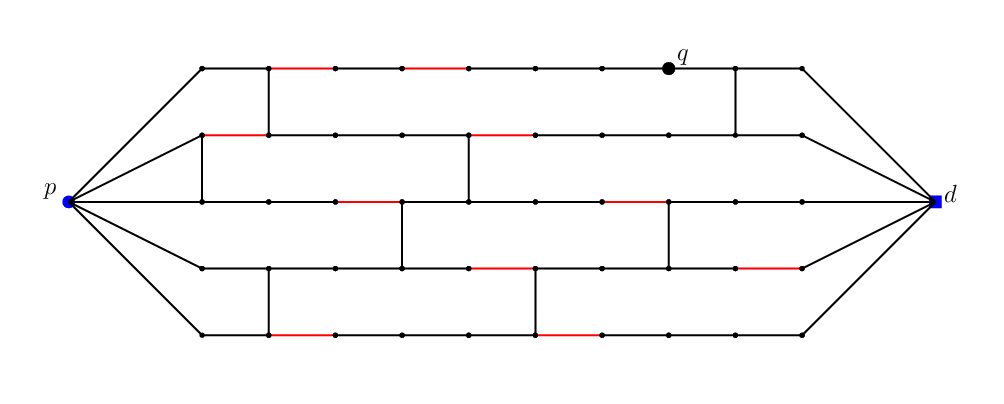}
\caption{Sample instance.}
\label{fig2} 
\end{center}
\end{figure}

After analyzing the previous instances, we identified that the naive method can perform as well as KSPP-RPP in a well-connected graph where the rerouting penalty is low i.e. UGV does not lose much time if it had to change its path, which is the case with uniform grid graphs. Therefore, we generated instances that would enforce a high rerouting cost for the UGV. An example of such an instance is shown in Figure \ref{fig2}. In this example, there are 5 different paths from $p$ to $d$ and they are only connected by a few bridges. The bridges are strategically placed so that the UGV has to incur a high cost if it has to change its path. For these instances, we force the UGV to change its initial path, which is based on the expected cost, by acting as an adversary and assigning the true cost as $T_e^{max}$ for all impeded edges on the initial path. The true cost of other impeded edges is set to $T_e^{min}$. For this study, we generated 100 similar instances with starting position ($p$) and destination ($d$) for the UGV set to $(0,0)$ and $(200,0)$ respectively, and with 10 different paths connecting $p$ and $d$. We randomly select two edges from each path to be impeded. $10\%$ of the nodes from each path are randomly selected to connect them with a neighboring path using a bridge. $q$ is chosen randomly for each instance. The position of all nodes, excluding $p$ and $d$, resembles a grid with the top left corner placed at $(10,100)$ and the bottom right corner at $(190, -90)$. We follow the same cost structure for the graph as described above. Computation results for this set of instances are shown in Table \ref{table1}. We ran the simulation with different $k$ values for the D*KSPP-RPP algorithm for the given set of instances.
In Table \ref{table1}, each row represents the average value of the Lower Bound ($LB$), cost from the naive method ($\hat{C}$), and cost from D*KSPP-RPP ($C$) over 100 instances. $\delta$ represents the improvement of D*KSPP-RPP over the naive method and is computed as follows: $(\hat{C}-C)/(\hat{C}-LB)\times100$. $\hat{\sigma}$ and $\sigma$ represent the standard deviation of cost for the naive method and D*KSPP-RPP respectively. As we can see, the proposed algorithm performs the same as the naive method for $k=1$ and it improves upon the naive method as we increase $k$. D*KSPP-RPP can improve upon the naive method up to $28\%$. Also, the standard deviation, $\sigma$, does not increase which indicates that the improvement is consistent across all instances. It is also important to note that the results do not improve as we increase $k$ beyond $5$. This is expected because as we increase $k$, we are increasing our search space which will force us to visit some impeded edges that are not relevant and delay the inspection of important edges.


\begin{table}[htbp]
\vspace*{0.1in}
\caption{D*KSPP-RPP vs Naive Method Results}
\label{table1}
    \begin{tabular}{c @{\hspace{0.8cm}} c @{\hspace{0.8cm}} c @{\hspace{0.8cm}} c @{\hspace{0.8cm}} c @{\hspace{0.8cm}} c @{\hspace{0.8cm}} c}
        \toprule
        $k$ & $LB$ & $\hat{C}$ & $C$ & $\delta(\%)$ & $\hat{\sigma}$ & $\sigma$ \\ 
        \cmidrule(lr){1-7}
        1  &  208 & 237.3 & 237.3  & 0.0   & 29.8 & 29.8 \\
        2  &  208 & 237.3 & 232.3  & 17.2  & 29.8 & 27.4 \\
        3  &  208 & 237.3 & 231.2  & 20.7  & 29.8 & 27.5 \\
        4  &  208 & 237.3 & 230.2  & 24.2  & 29.8 & 27.7 \\
        5  &  208 & 237.3 & 229.1  & 27.9  & 29.8 & 27.8 \\
        6  &  208 & 237.3 & 229.7  & 26.0  & 29.8 & 28.4 \\  
        7  &  208 & 237.3 & 222.6  & 26.3  & 29.8 & 28.7 \\  
        \bottomrule
    \end{tabular}
\end{table}

\subsection{Comparison between RPP and PAA for the UAV planning} \label{RPP-PAA}

For the following analysis, we set the maximum computation time limit for replanning to 1 second. This can change based on different applications. 
As shown in \cite{monroy2014rural}, the computation time to solve an RPP instance increases dramatically as we increase the number of edges to be visited and we observe a similar behavior in our implementation. First, we performed several simulations to fine-tune the parameters for the PAA and the best results were observed at $(p_1,\,p_2,\,p_3,\,p_4) = (0.25,\,0.25,\,0.2,\,0.3)$.
For this study, we generated 100 random instances with the same structure as in the previous section. For each instance, we randomly pick $20\%$ of edges in each path to be impeded. Also, we randomly pick $20\%$ of nodes from each path to connect to a neighboring path. We follow the same cost structure as defined in Section \ref{KSPP-Naive} and the true cost of impeded edges is realized from the associated distribution. 

In Tabel \ref{table2}, $C_1$ and $C_2$ represent the average value of the cost from D*KSPP-RPP and D*KSPP-PAA algorithms for 100 instances. $LB$ and $\hat{C}$ are same as before. $t_1$ and $t_2$ represent the maximum computation time for RPP and PAA respectively across 100 instances. We performed the simulation with different $k$ values for D*KSPP because as $k$ increases, the number of critical edges considered by the UAV also increases.
As shown in Table \ref{table2}, the performance of PAA is very close to RPP. The computation time for RPP increases significantly with $k$ as expected. However, the computation time for PAA remains lower than RPP by several orders of magnitude and only increases marginally with $k$.

\begin{table}[htbp]
\caption{Comparison between the RPP vs PAA algorithms for the UAV}
\label{table2}
    \begin{tabular}{c @{\hspace{0.8cm}} c @{\hspace{0.8cm}} c @{\hspace{0.8cm}} c @{\hspace{0.8cm}} c @{\hspace{0.8cm}} c @{\hspace{0.8cm}} c}
        \toprule
        $k$ & $LB$ & $\hat{C}$ & $C_1$ & $C_2$ & $t_1$(s) & $t_2$(s) \\ 
        \cmidrule(lr){1-7}
        1  &  265.4 & 280.4 & 280.3  & 280.3  & 1.8e-4 & 4.4e-5 \\
        2  &  265.4 & 280.4 & 279.1  & 279.1  & 7.1e-3 & 6.8e-5\\
        3  &  265.4 & 280.4 & 277.4  & 277.6  & 0.99   & 7.8e-5 \\
        4  &  265.4 & 280.4 & 277.0  & 278.1  & 2.85   & 9.2e-5\\
        5  &  265.4 & 280.4 & 277.9  & 277.8  & 118    & 8.6e-5\\
        
        \bottomrule
    \end{tabular}
    
\end{table}

\subsection{Limitations of D*KSPP} \label{Limit KSPP}



\begin{figure}[htbp]
\begin{center}
\includegraphics[width=\linewidth]{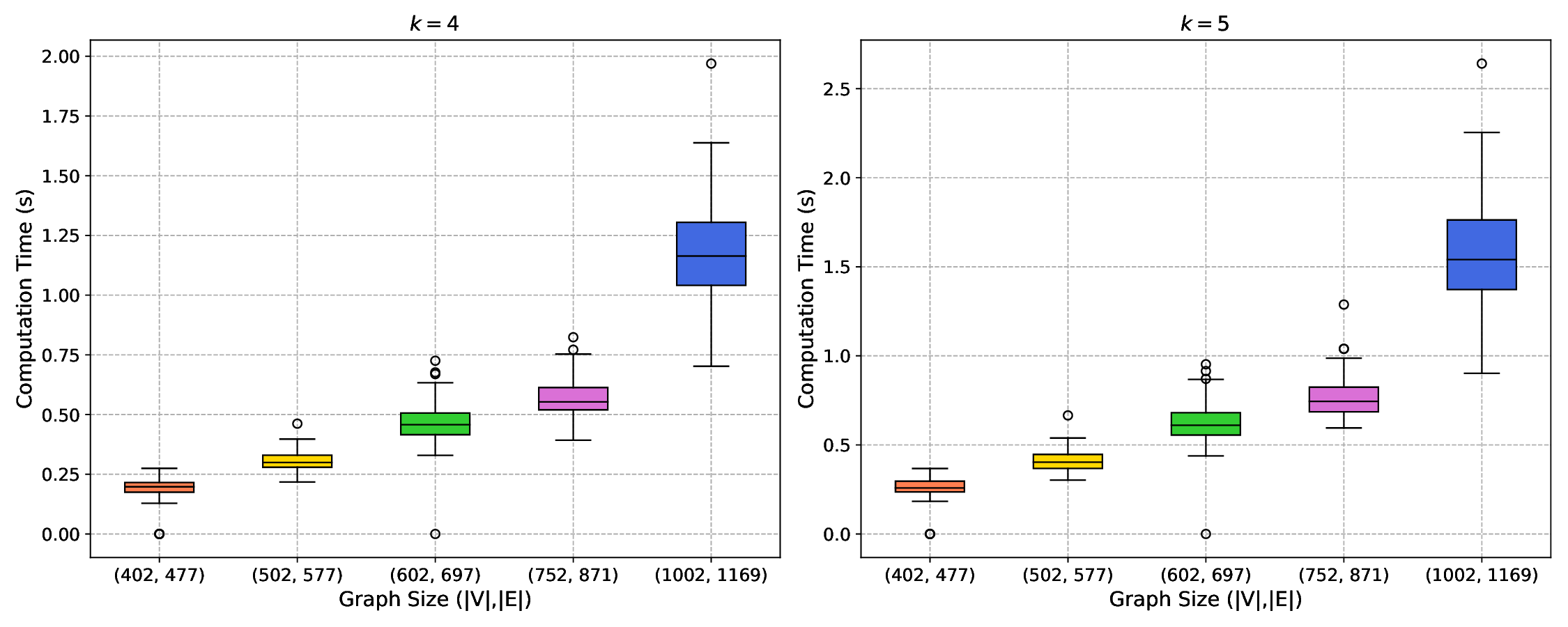}
\caption{Computation time of D*KSPP with $k$ set to $4$ and  $5$ for different graph sizes.}
\label{KSPP_all} 
\end{center}
\end{figure}

In this section, we explore the limitations of D*KSPP. We want to understand the impact of graph size on the replanning computation for the UGV and the maximum size of graph D*KSPP can handle without exceeding the computation time limit of 1 second. For this study, we generated similar instances as Figure \ref{fig2}. As the nodes follow a grid pattern except for $p$ and $d$, we progressively increase the grid size as follows: $20\times20$, $25\times20$, $30\times20$, $30\times25$, $40\times25$. This results in instance with number of nodes, $|V|$, ranging from 400 to 1000. For each graph size, we generated 100 instances. For each instance, we randomly pick $30\%$ of the edges on each path to be impeded and $30\%$ of the nodes on each path are connected to the neighboring path with a bridge. We follow the same cost structure defined in section \ref{KSPP-Naive} and the UAV's starting position is chosen randomly for each instance. Figures \ref{KSPP_all} represents the distribution of maximum computation for UGV replanning with $k$ set to 4 and 5. It is shown that for $k=4$, the maximum computation time does not exceed our limit of 1 second for all instances except for last set which has about $1000$ nodes. For $k=5$, a few instances with $(|V|,|E|) = (752,871)$ exceed the time limit. For graphs with about $1000$ nodes, it is difficult to contain the computation time below 1 second. However, for most practical applications, an environment can easily be described using $1000$ or less node. Additionally, we can always reduce the larger graph by various pre-processing techniques to eliminate redundant nodes, e.g. removing nodes with degree 2 that are not incident on any impeded edges.

\subsection{Real-World Case Study} \label{RoadNetworkResults}


\begin{figure}[] 
  \centering
  \includegraphics[width=\linewidth]{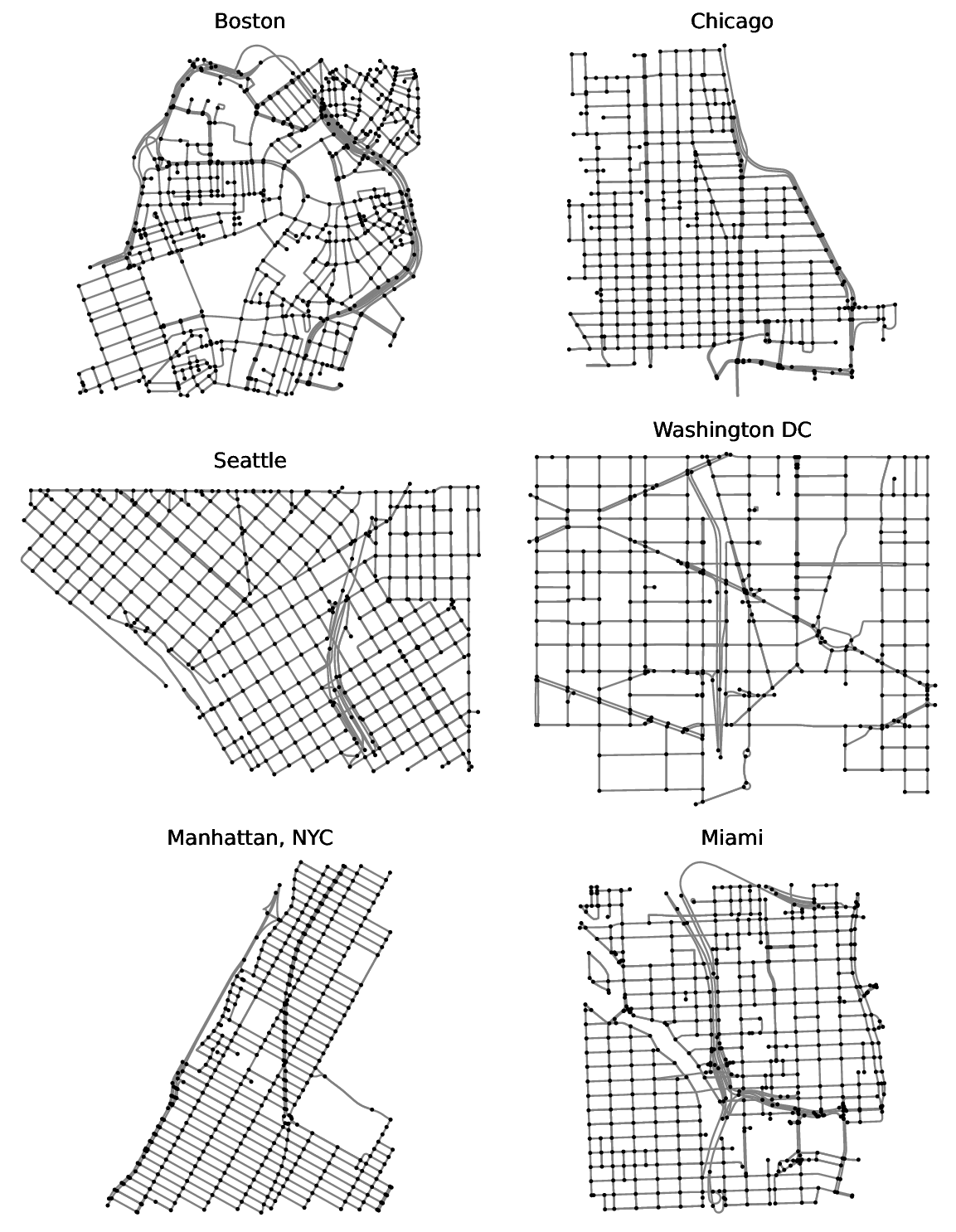}
  \caption{OpenStreetMap\cite{OpenStreetMap} road networks from six major cities in the USA.}
  \label{allcities}
\end{figure}

\begin{table}[]
    \caption{Real-World Case Study}\label{tab:RoadNetworkResults}
    \begin{tabular*}{\textwidth}{@{\extracolsep\fill}l c c c c c c}
        \toprule
        \multirow{2}{*}{\textbf{City}} & & \multicolumn{5}{c}{\textbf{$|K|/|E|$}} \\
        \cmidrule{3-7}
        & & 0.4 & 0.5 & 0.6 & 0.7 & 0.8\\
        \midrule
        \multirow{4}{*}{Boston, MA} & $LB$ & 4726.1 & 5670.1 & 6783.8 & 8087.3 & 10016.4 \\
        & $\hat{C}$ & 5047.8 & 6202.5 & 7586.2 & 9124.9 & 11377.7 \\
        & $C$ & 5021.9 & 6152.3 & 7505.6 & 8983.9 & 11224.1 \\
        & $\delta$ & 8.04 & 9.43 & 10.05 & 13.58 & 11.28 \\
        \midrule
        \multirow{4}{*}{Chicago, IL} & $LB$ & 5470.8 & 6492.1 & 8218.7 & 9179.49 & 11120.7 \\
        & $\hat{C}$ & 5828.2 & 6955.5 & 8944.0 & 10265.1 & 12648.8 \\
        & $C$ & 5805.3 & 6930.4 & 8896.7 & 10067.3 & 12378.6 \\
        & $\delta$ & 6.42 & 5.41 & 6.52 & 18.21 & 17.68 \\
        \midrule
        \multirow{4}{*}{Seattle, WA} & $LB$ & 3297.3 & 3803.2 & 4666.4 & 5720.8 & 7041.3 \\
        & $\hat{C}$ & 3523.6 & 4212.5 & 5364.4 & 6587.2 & 8173.5 \\
        & $C$ & 3515.4 & 4198.0 & 5306.9 & 6483.0 & 7977.3 \\
        & $\delta$ & 3.62 & 3.53 & 8.24 & 12.02 & 17.32 \\
        \midrule
        \multirow{4}{*}{Washington DC} & $LB$ & 3801.1 & 4494.9 & 5293.5 & 6439.6 & 7840.3 \\
        & $\hat{C}$ & 4023.1 & 4893.4 & 5864.0 & 7340.8 & 9053.2 \\
        & $C$ & 4008.9 & 4874.9 & 5793.7 & 7239.1 & 8871.6 \\
        & $\delta$ & 6.41 & 4.63 & 12.32 & 11.29 & 14.97 \\
        \midrule
        \multirow{4}{*}{Manhattan, NYC} & $LB$ & 6726.9 & 7719.9 & 9037.2 & 10837.3 & 12483.2 \\
        & $\hat{C}$ & 7038.0 & 8177.1 & 9682.6 & 11814.6 & 14061.2 \\
        & $C$ & 7002.1 & 8133.3 & 9602.8 & 11708.1 & 13906.7 \\
        & $\delta$ & 11.55 & 9.58 & 12.36 & 10.90 & 9.79 \\
        \midrule
        \multirow{4}{*}{Miami, FL} & $LB$ & 5058.9 & 5949.6 & 7110.6 & 8749.2 & 10372.7 \\
        & $\hat{C}$ & 5378.8 & 6485.8 & 7984.6 & 9882.4 & 11774.4 \\
        & $C$ & 5365.1 & 6455.5 & 7924.5 & 9735.1 & 11637.7 \\
        & $\delta$ & 4.27 & 5.64 & 6.88 & 12.98 & 9.75 \\
        \bottomrule
    \end{tabular*}
\end{table}

In this section, we evaluate the performance of our algorithm for real-world instances. For this study, we took the road networks from six different cities in the United States from OpenStreetMap\cite{OpenStreetMap}. An example of these road networks is shown in Figure \ref{allcities}.
For each road network, we varied the number of impeded edges in the network to represent different levels of traffic congestion. The fraction of impeded edges, $|K|/|E|$, is varied from $0.4$ to $0.8$. For a given value of fraction of impeded edges, we generate $100$ different instances where we randomly select the impeded edges. $T_e$ for all unimpeded edges and $T_e^{min}$ for all impeded edges is set to the distance of the edge provided by OpenStreetMap. For all impeded edges, $T_e^{max}$ is set to be 10 times the distance of that edge. The distance of an edge between the nodes may not be equal to the euclidean distance between the nodes as the roads can be winding in nature. Therefore, the UAV may travel along an edge for inspection or it can move in a straight line to reach the next node. The travel cost of the UAV between two nodes is then the distance traversed by the UAV divided by the UAV speed. The starting position and destination of the UGV are chosen to be two nodes that are farthest from each other in the graph. The starting position of the UAV is chosen randomly for each instance. In this analysis, we have set $k=3$ for D*KSPP and used PAA for UAV path planning. 

The results are shown in Table \ref{tab:RoadNetworkResults}. $LB$, $\hat{C}$, and $C$ represent the average values of Lower Bound, cost from naive method, and cost from D*KSPP-PAA over 100 instances. $\delta$ is again the improvement of our proposed algorithm over the naive method and is same as defined in Section \ref{KSPP-Naive}. We can see that the proposed algorithm performs better than the naive method for all case studies. We also see an improvement in performance as we increase the fraction of impeded edges in the graph for most of the cities. This indicates that our approach gets more effective as we increase the traffic congestion.

\section{Conclusion} \label{conclusion}
We have considered the problem of a UGV traveling through an environment to reach a desired destination and being assisted by a UAV. The environment has certain obstructions that can slow down the progress of the UGV and the severity of these obstructions is unknown. We took an example of a road network as the environment where few roads are affected by traffic congestion. The travel time/cost for UGV through these impeded paths is a random variable. The UAV can assist the UGV by realizing the true travel cost of these impeded paths in advance so that the UGV can reroute its path if needed. We have presented an online algorithm to solve the problems in real time for time-critical applications where we do not have enough time for offline planning. We present a Dynamic $k$-Shortest Path Planning (D*KSPP) algorithm for UGV path planning and a Rural Postman Problem with Time Window (RPP-TW) algorithm for UAV planning. We also present a heuristic-based priority assignment algorithm (PAA) for the UAV to solve large instances. We then present extensive computational results to corroborate the algorithms along with real-world analysis using road networks from six major cities in the USA. Future work can include extending the presented problem to have a group of UGVs and UAVs working together such that UGVs can reach their respective destination in minimum time. 

\break





\begin{appendices}

\section{D* Lite Methods}\label{AppA}
\begin{algorithm}
        \caption{D* Lite Methods \label{D* methods}}
        \begin{algorithmic}[1]
           
            \Procedure{CalculateKey}{$v$}
                \State return [$min(g(v), rhs(v)) + h(v, v_{curr}) + k_m; min(g(v), rhs(v))$]
            \EndProcedure

            \vspace{2mm}
            
            \Procedure{UpdateVertex}{$v$}
                \If{$g(v)\neq rhs(v)$ AND $v\in Q$} Q.Update($v$, CalculateKey($v$))
                
                \ElsIf{$g(v)\neq rhs(v)$ AND $v\notin Q$} Q.Insert($v$, CalculateKey($v$))
             
                \ElsIf{$g(v)=rhs(v)$ AND $v\in Q$} Q.Remove($v$)
                \EndIf
            \EndProcedure

            \vspace{2mm}
            
            \Procedure{ComputeShortestPath}{}
                \While{Q.TopKey() $<$ Calculate($v_{curr}$) OR $rhs(v_{curr}) > g(v_{curr})$}
                    \State $v \gets $ Q.Top()
                    \State $k_{old} \gets $ Q.TopKey()
                    \State $k_{new}  \gets $ CalculateKey($v$)
                    \If{$k_{old} < k_{new}$} Q.Update($v, k_{new}$)
                    \ElsIf{$g(v)>rhs(v)$}
                        \State $g(v) = rhs(v)$
                        \State Q.Remove($v$)
                        \For{$s\in N(v)$}
                            \If{$s\neq d$} $rhs(s) = min(rhs(s), c(s,v)+g(v))$
                            \EndIf
                            \State UpdateVertex($s$)
                        \EndFor
                    \Else{}
                        \State $g_{old} \gets g(v)$
                        \State $g(v) \gets \infty$
                        \For{$s\in N(v)\cup {u}$}
                            \If{$rhs(s) = c(s,v)+g_{old}$}
                                \If{$s\neq d$} $rhs(s) \gets min_{s'\in N(s)}(c(s,s')+g(s'))$
                                \EndIf
                            \EndIf
                            \State UpdateVertex($s$)
                        \EndFor
                    \EndIf
                \EndWhile
            \EndProcedure

            \vspace{2mm}

        \end{algorithmic}
\end{algorithm}





\end{appendices}


\bibliography{sn-bibliography}

\end{document}